\documentclass[letterpaper]{article} 
\usepackage{aaai25}  
\usepackage{times}  
\usepackage{helvet}  
\usepackage{courier}  
\usepackage{cite}
\usepackage[hyphens]{url}  
\usepackage{graphicx} 
\usepackage{natbib}  
\usepackage{caption} 
\usepackage{algorithm}
\usepackage{algorithmic}
\usepackage{svg}
\usepackage{newfloat}
\usepackage{listings}
\DeclareCaptionStyle{ruled}{labelfont=normalfont,labelsep=colon,strut=off} 
\floatstyle{ruled}
\newfloat{listing}{tb}{lst}{}
\floatname{listing}{Listing}
\title{StegaVision: Enhancing Steganography with Attention Mechanism

(Student Abstract)}
\author{
   Abhinav Kumar\textsuperscript{\rm 1}\equalcontrib, 
   Pratham Singla\textsuperscript{\rm 1}\equalcontrib, 
   Aayan Yadav\textsuperscript{\rm 1} }

\affiliations{
    \textsuperscript{\rm 1}Vision and Language Group, Indian Institute of Technology Roorkee, Roorkee, Uttarakhand, India - 247667 \\
   abhinav\_k@ma.iitr.ac.in , pratham\_s@me.iitr.ac.in, aayan\_y@mfs.iitr.ac.in
}

\begin{document}

\maketitle

\begin{abstract}
Image steganography is the technique of embedding secret information within images. The development of deep learning has led to significant advances in this field. However, existing methods often struggle to balance image quality, embedding capacity, and security. This paper proposes a novel approach to image steganography by enhancing an encoder-decoder architecture with attention mechanisms, specifically focusing on channel and spatial attention modules. We systematically investigate five configurations: (1) channel attention, (2) spatial attention, (3) sequential channel followed by spatial attention, (4) spatial attention followed by channel attention and (5) parallel channel and spatial attention. Our experiments show that adding attention mechanisms improves the ability to embed hidden information while maintaining the visual quality of the images. The increase in the PSNR and SSIM scores shows that using a parallel combination of channel and spatial attention improves image quality and hiding capacity simultaneously. This is in contrast to previous works where there is a tradeoff between them. This study shows that attention mechanisms in image steganography lead to better hiding of secret information. Our code is available at \url{https://github.com/vlgiitr/StegaVision}.

\end{abstract}
\begin{figure*}[h]
\centering
\includegraphics[width=0.95\linewidth, height=0.23\textheight]{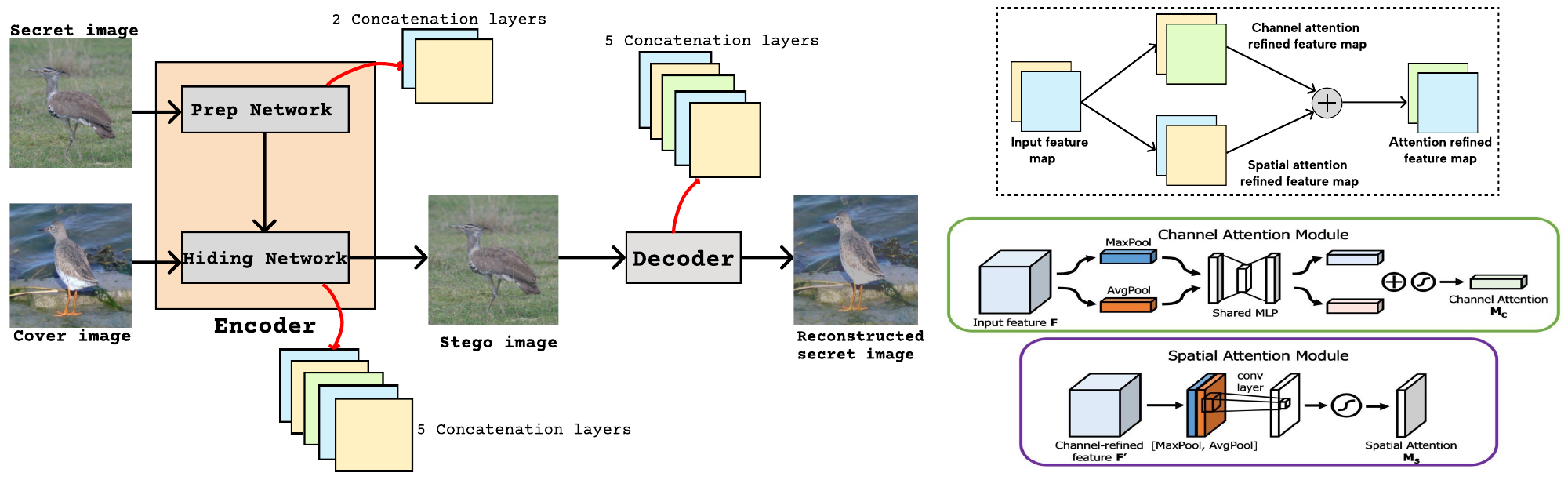} 
\caption{\textbf{Left: }Encoder-decoder-based steganography model \cite{baluja2017hiding}. \textbf{Right Top: }Application of channel and spatial attention in parallel. \textbf{Right Bottom: }Calculation of channel and spatial attention. }
\label{approach}
\end{figure*}

\section{Introduction}{
Steganography is the practice of hiding information within other data. This has evolved significantly with deep learning techniques. While traditional methods like Least Significant Bit (LSB) and transform domain techniques have laid the foundation, they often suffer from limitations in robustness and capacity. Deep learning-based approaches, particularly convolutional neural networks (CNNs) \cite{subramanian2021image}, have shown promise in addressing these challenges by learning complex patterns for embedding and extracting hidden data. The attention mechanism has recently been applied to steganography to enhance the performance of embedding and extraction processes in GAN-based architectures \cite{tan2021channel}. Channel attention focuses on \emph{what} features are important by assigning different weights to each channel. This allows the model to emphasize informative features and suppress irrelevant ones. Spatial attention, conversely, determines \emph{where} to focus within the feature maps, refining the spatial details critical for the task of image steganography. 
 \newline
Our study examines the effectiveness of integrating channel and spatial attention mechanisms within encoder-decoder-based models for steganography. We experiment with different configurations of attention mechanisms to determine how these modifications impact the quality and security of steganography. We hope the insights we provide will guide future research in this field.
 
}

\begin{table*}[!t]
\centering      
\resizebox{\textwidth}{!}{%
\begin{tabular}{|c|c|c|c|c|c|c|}
\hline
\textbf{Model} & \textbf{PSNR Cover} & \textbf{SSIM Cover} & \textbf{PSNR Secret} & \textbf{SSIM Secret} & \textbf{MSE Loss Cover} & \textbf{MSE Loss Secret} \\ \hline
\textbf{Baseline} & 10.658 & 0.831 & 10.276 & 0.796 & 0.1620 & 0.1701 \\ \hline
\textbf{Channel Only} & 10.666 & 0.831 & \textit{10.289} & 0.797 & 0.1619 & \textit{0.1699} \\ \hline
\textbf{Spatial Only} & \textbf{10.734} & \textbf{0.835} & 10.266 & \textit{0.799} & \textbf{0.1612} & 0.1703 \\ \hline
\textbf{Channel-Spatial Parallel} & \textit{10.672} & \textit{0.832} & \textbf{10.431} & \textbf{0.808} & \textit{0.1616} & \textbf{0.1684} \\ \hline
\textbf{Channel then Spatial} & 10.504 & 0.815 & 10.090 & 0.779 & 0.1640 & 0.1725 \\ \hline
\textbf{Spatial then Channel} & 10.614 & 0.827 & 10.172 & 0.787 & 0.1627 & 0.1715 \\ \hline
\end{tabular}%
}
\caption{Performance comparison of different attention mechanism configurations in steganography models. 
Best results are in \textbf{bold} and second best results are in \textit{italics}.}
\label{tab:attention_results}
\end{table*}

\section{Methodology}{

We examine the impact of integrating attention mechanisms into an encoder-decoder-based steganography model. We experiment with five configurations involving channel and spatial attention mechanisms \cite{woo2018cbam}. These configurations were designed to evaluate how these attention mechanisms influence the model's performance in hiding and reconstructing secret images within cover images.

Encoder-decoder architecture from \citet{baluja2017hiding} is the baseline model. The encoder consists of two networks: one is a prep network, and the other is a hiding network. The prep network takes the secret image and, after 2 convolutional blocks, converts it from a 3-channel feature map to a 65-channel feature map and concatenates it with the cover image. This is passed as input to the hiding network, which consists of 5 more such blocks. The final output of the encoder block is a stego image. The stego image is passed to a decoder whose architecture is similar to the hiding network. The output of the decoder is a reconstructed secret image.

We apply different forms of attention between two convolutional blocks. Each convolution block consists of filters of sizes 3, 4, and 5 that extract 50, 10, and 5 feature maps, respectively. These feature maps are then concatenated and passed to the next convolution block. In the baseline, normal concatenation is used, while in channel attention, the concatenated feature map is reweighted using channel attention to see which feature map is relatively more important. The spatial attention map reweights each feature map to see which region is more important in each feature map.

We use MSE, PSNR, and SSIM scores to measure pixel-wise distortion between images, the image quality of images, and structural similarity between images, respectively.

\section{Results} {

Table 1 compares the performance metrics across various configurations of attention mechanisms in the steganography model. The results demonstrate that the parallel integration of channel and spatial attention outperforms other configurations. It offers the best trade-off between maintaining high image quality and hiding capacity, as reflected in the superior PSNR and SSIM scores. When applied individually, channel and spatial attention also enhance the model, but to a lesser extent. Sequential configurations, particularly channel-then-spatial, underperform, highlighting that the order and method of attention integration are crucial for optimizing the effectiveness of steganography. Overall, attention mechanisms are valuable additions for enhancing encoder-decoder-based image steganography.

\section{Conclusion} {
This paper demonstrates that integrating channel and spatial attention mechanisms into the encoder-decoder architecture enhances the performance of steganography.  We evaluate five different configurations of attention modules and find that using channel and spatial attention together, especially in parallel, leads to better results than the baseline model.  This highlights that attention mechanisms improve the feature representation for hiding images.  Our results suggest that attention-based models provide an effective approach for achieving secure, high-quality image steganography.
}
\section{Future Scope} {
There is significant potential for further research and experimentation to strengthen and enhance the performance and applicability of these techniques. Future work could involve testing the approach on larger and more advanced models. We could also use multi-scale attention in different layers of the encoder-decoder model to better capture spatial and channel-wise features. To get a clearer idea of how well the model performs, it is important to test it with real-world images and a variety of datasets. This will help address the current limitations and enhance the security and efficiency of steganographic models. 

}

\bibliography{aaai25} 

\end{document}